\documentclass[conference]{IEEEtran}
\IEEEoverridecommandlockouts
\usepackage{cite}
\usepackage{amsmath,amssymb,amsfonts}
\usepackage{algorithmic}
\usepackage{graphicx}
\usepackage{subfigure}
\usepackage{textcomp}
\usepackage{url}
\usepackage{xcolor}
\def\BibTeX{{\rm B\kern-.05em{\sc i\kern-.025em b}\kern-.08em
    T\kern-.1667em\lower.7ex\hbox{E}\kern-.125emX}}
\begin{document}

\title{Segmentation of Skeletal Muscle in Thigh Dixon MRI Based on Texture Analysis
\thanks{This research was funded by Funda\c{c}\~ao para a Ci\^encia e Tecnologia, under the research grant SFRH/BD/130858/2017, and the Instituto de Telecomuni\-ca\c{c}\~oes - Funda\c{c}\~ao para a Ci\^encia e Tecnologia, under the project UID/EEA/50008/2019.}
}

\author{\IEEEauthorblockN{Rafael Rodrigues and Antonio M. G. Pinheiro}
\IEEEauthorblockA{\textit{Instituto de Telecomunica\c{c}\~oes} \\
\textit{Universidade da Beira Interior}\\
Covilh\~a, Portugal \\
rafael.rodrigues@ubi.pt, pinheiro@ubi.pt}
}

\maketitle

\begin{abstract}
Segmentation of skeletal muscles in Magnetic Resonance Images (MRI) is essential for the study of muscle physiology and diagnosis of muscular pathologies. However, manual segmentation of large MRI volumes is a time-consuming task. The state-of-the-art on algorithms for muscle segmentation in MRI is still not very extensive and is somewhat database-dependent. In this paper, an automated segmentation method based on AdaBoost classification of local texture features is presented. The texture descriptor consists of the Histogram of Oriented Gradients (HOG), Wavelet-based features, and a set of statistical measures computed from both the original and the Laplacian of Gaussian filtering of the grayscale MRI. The classifier performance suggests that texture analysis may be a helpful tool for designing a generalized and automated MRI muscle segmentation framework. Furthermore, an atlas-based approach to individual muscle segmentation is also described in this paper. The atlas is obtained by overlaying the muscle segmentation ground truth, provided by a radiologist, after image alignment using an appropriate affine transformation. Then, it is used to define the muscle labels upon the AdaBoost binary segmentation. The developed atlas method provides reasonable results when an accurate muscle tissue segmentation was obtained.
\end{abstract}


\section{Introduction}
\label{sec:intro}
Muscle segmentation in medical imaging enables quantitative measurements of muscle tissue and fat infiltration. Such outcomes are crucial to most physiological and/or pathological evaluations. Manual segmentation of muscles in Magnetic Resonance Images (MRI) can be an extremely arduous task, but it is still a common approach in radiology services. Moreover, results are often affected by intra- and inter-operator variability. An automated framework would increase the ability to deal with large Whole Body MRI datasets.

Some efforts have recently been made towards the development of segmentation algorithms for skeletal muscle in MRI. Most of the approaches are atlas-based, such as those presented in \cite{Ahmad2014, Sdika2016, Troter2016, karlsson2015automatic}. Despite some interesting results, relying on a segmentation exclusively based on a muscle atlas may have some disadvantages, regarding the availability of annotated databases and the variability on muscle geometry, depending on the body region and, to a lesser extent, the subject.

Orgiu \emph{et al.} \cite{orgiu2016automatic} developed a framework for muscle and fat segmentation in \emph{T1-weighted} MRI, using fuzzy \emph{C-means} clustering and an active contour (snake) approach. The reported method provided accurate segmentation results with a fairly simple approach. Nonetheless, muscle signal intensity in \emph{T1w} MRI is visually quite distinctive, and reasonably easy to segment using intensity-based clustering methods, as well as snakes \cite{rodrigues2015two}. However, these conditions may not be verified with different MR imaging sequences. 

Other proposed methods include a graph-based approach, using the Random Walks algorithm developed by Baudin et al. \cite{Baudin2012}, and a probabilistic model using Gabor Features, which is presented in \cite{Andrews2015}. 

Given the current limitation of research on the topic, considering \emph{2D} MRI slices or \emph{3D} MRI volumes, automated muscle segmentation leads to a challenging task, which could benefit largely from a more generalized approach. In this paper, a fully automated method for the segmentation of skeletal muscle in Dixon MRI scans of the human thigh is proposed. In \cite{Andrews2015}, the authors studied \emph{T1w} MR images of the human thigh and discuss that thigh muscles do not show discernible variations in texture between each other. Nonetheless, it is possible that muscle present intrinsic texture patterns that are invariably distinct from other tissues in the thigh. Based on this assumption, our research studies the feasibility and effectiveness of texture analysis in discriminating muscle from other tissues. The proposed segmentation method relies on supervised learning using texture-based features. An atlas is used to assign muscle labels to the classifier result.

\begin{figure*}[t!]
	\begin{center}
	\subfigure[Original Dixon MRI]{
			\includegraphics[width=0.23\linewidth]{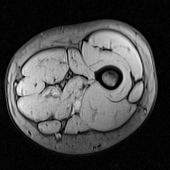}
		}
	\subfigure[Full HOG representation]{
			\includegraphics[width=0.23\linewidth]{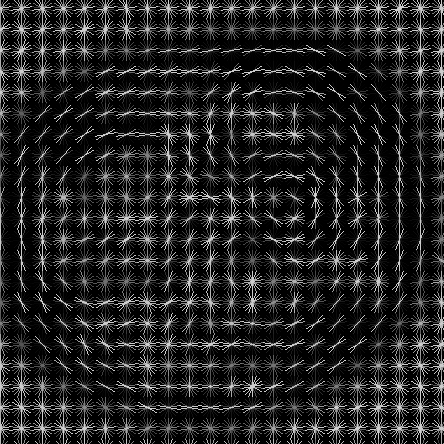}
		}
	\subfigure[LoG filtered image ($\sigma = 1.5$)]{
			\includegraphics[width=0.23\linewidth]{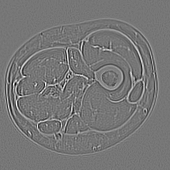}
		}
	\subfigure[\emph{Wavelet} \emph{low pass} coefficients]{
			\includegraphics[width=0.23\linewidth]{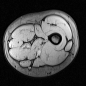}
		}\\
	\subfigure[\emph{Wavelet} horizontal detail coefficients]{
			\includegraphics[width=0.23\linewidth]{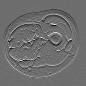}
		}
	\subfigure[\emph{Wavelet} vertical detail coefficients]{
			\includegraphics[width=0.23\linewidth]{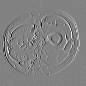}
		}
	\subfigure[\emph{Wavelet} diagonal detail coefficients]{
			\includegraphics[width=0.23\linewidth]{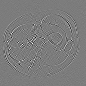}
		}
	\subfigure[Ground truth of skeletal muscles after erosion]{
			\includegraphics[width=0.23\linewidth]{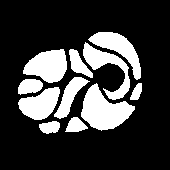}
		}		
	\caption{Out-of-phase Dixon MRI scan (a). Image preprocessing results for texture feature extraction (b) to (g). Muscle ground truth (h)}
		\label{fig:desc}
	\end{center}
\end{figure*}

\section{Materials and Methods}
\label{sec:methods}
A dataset comprising 10 out-of-phase 3pt Dixon MRI volumes was used for algorithm development and testing. Images were acquired using a 3T Siemens scanner, with a spatial resolution of $1mm \times 1mm \times 5mm$, and the following parametrization: $TR = 10ms$, $TE_{1} = 2.75ms$, $TE_{2} = 3.95ms$, $TE_{3} = 5.15 ms$, RF flip angle $= 3°$.

Manual segmentation of skeletal muscles was performed by an expert radiologist and used ground truth for classifier training and performance assessment.

Each volume contained 80 scans of the right thigh (Fig. \ref{fig:desc}(a)). MRI slice size is 256x256 pixels. To avoid including slices with too few muscle tissue, slices near the knee and the ankle were discarded \emph{a priori}, resulting in a subset of 40 MRI slices. From this initial selection, 5 consecutive slices were randomly selected from each volume, to simultaneously enable the proposed muscle atlas construction approach and address memory limitations of the experimental setup.

\subsection{Feature extraction}
\label{sec:feature}
Images are subdivided into a 16x16 non-overlapping grids for local texture description. Each 16x16 pixel window is primarily represented by a set of coarser resolution texture features, including the Histogram of Orientated Gradients (HOG) \cite{HOG2005} and statistical measures from both the original grayscale image and a Laplacian of Gaussian (LoG) filtered image \cite{LOG2009}. For a finer resolution of the descriptor, Wavelet coefficients \cite{Zhang2011, EL2010} from a 3-level decomposition were also included, resulting in a 54-bin descriptor.

\subsubsection{Histogram of Oriented Gradients}
\label{sec:hog}
The first well-known application of HOG was proposed by Dalal and Triggs \cite{HOG2005}, who used the descriptor for pedestrian detection in still images. This descriptor provides a gridded representation of image gradients, in terms of magnitude and orientation.

In this paper, the Dalal-Triggs variant was applied, using the \emph{VLFeat} MATLAB implementation \cite{vedaldi08vlfeat}, with 16x16 blocks and considering 9 orientations, resulting in a 36 bin histogram. An example of the full HOG representation of a MRI scan is shown in Fig. \ref{fig:desc}(b).

\subsubsection{Statistical features}
\label{sec:statistical}
Obtaining local statistical measures of grayscale images may further provide a description of local texture by describing numerically the local intensity variations. In this research, both the original MRI grayscale image (Fig. \ref{fig:desc}(a)) and the outcome of a 5x5 LoG filter with $\sigma = 1.5$ \cite{LOG2009} (Fig. \ref{fig:desc}(c)) are used for statistical feature extraction.

While the original grayscale image provides important differences in intensity and its variations inside tissues, the LoG filter allows for edges highlighting. Those two images are subdivided into 16x16 blocks and the mean, variance, skewness and kurtosis are computed computed inside each block.

\subsubsection{Wavelet coefficients}
\label{sec:wavelet}
The Discrete Wavelet Transform (DWT) was applied with the Haar wavelet \cite{Zhang2011, EL2010}. A 3-level decomposition was considered, resulting in a 10-bin descriptor, including the \emph{low-pass} coefficients (Fig. \ref{fig:desc}(d)) after the 3 decompositions and the detail coefficients of each level, e.g., horizontal (Fig. \ref{fig:desc}(e)), vertical (Fig. \ref{fig:desc}(f)) and diagonal Fig. \ref{fig:desc}(g)).

The resulting features provide a multi-resolution texture description. As image resolution is reduced to one half with each transform, each coefficient at level 1 corresponds to a 4-pixel neighborhood in the original image (2x2), at level 2 to a 16-pixel neighborhood (4x4), and at the coarser resolution to a 64-pixel neighborhood (8x8). The information contained in the \emph{wavelet} features complements the local descriptor, allowing to obtain some finer details in the segmentation contours.

\subsection{AdaBoost training and classification}
\label{sec:classification}
Texture features were classified using an AdaBoost classifier \cite{Ada1996, adaviola} with 500 iterations.
Manual segmentation of lower leg muscles, performed by a radiologist, was available for the entire set of MRI volumes. Using these masks as ground truth, binary masks were designed to set the training data labels as muscle / non-muscle. For each slice, every individual muscle region in the ground truth masks was eroded, as shown in Fig. \ref{fig:desc}(h). This step aims at avoiding an eventual inclusion of undesired intermuscular texture patterns into the positive training dataset.

To obtain the final label mask for AdaBoost training, the dominant label inside each 16x16 non-overlapping block is considered. In case of a tie, which would most likely occur in a region near non-muscle tissue, a negative label is assumed to further reduce the inclusion of non-muscle texture into the positive training set. Algorithm validation was done with a cross-validation setup, in which the training data for classifying 5 selected slices from each volume is extracted from the remaining 9 MRI volumes.

\begin{figure}[t!]
	\begin{center}
	\subfigure[\emph{Vastus medialis}]{
			\includegraphics[width=0.46\linewidth]{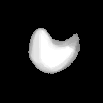}
		}
	\subfigure[\emph{Vastus intermedius}]{
			\includegraphics[width=0.46\linewidth]{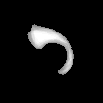}
		}\\
	\subfigure[\emph{Vastus lateralis}]{
			\includegraphics[width=0.46\linewidth]{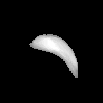}
		}
	\subfigure[Full muscle atlas]{
			\includegraphics[width=0.46\linewidth]{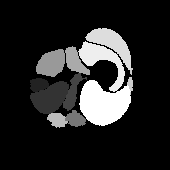}
		}
	\caption{Muscle atlas. From (a) to (c), the overlapping maps of three different muscles are presented. In (d), an example of a full atlas is shown.}
		\label{fig:atlas}
	\end{center}
\end{figure}

\subsection{Muscle labeling}
\label{sec:label}
A muscle atlas was defined using the same images involved in each cross-validation iteration. First, all images are registered to a common reference, selected randomly among the training set. Image registration relies on the convex hull of full set of muscle labels for each MRI slice and 2 key points - the bone centroid and the most distal point on the referred convex hull, with respect to the bone centroid. 


One image is randomly selected as a reference for the atlas relative positioning among the training set. In the first step, bone centroids of reference and target MRI slices are aligned through translation.

Considering the translated images, let $\vec{D_{1}}$ and $\vec{D_{2}}$ be the vectors connecting the bone centroid to the most distal point along the convex hull of the muscle labels, in the reference and target images, respectively. Images are then further aligned using rotation by the angle between vectors $\vec{D_{1}}$ and $\vec{D_{2}}$. Finally, considering the convex hull of the muscle labels in the reference image, the target images are scaled to match the reference.

After aligning the images within a common frame, a probability map is obtained for each muscle, overlapping the binary masks of all manually designed ROI with the same label (Figs. \ref{fig:atlas}(a), \ref{fig:atlas}(b) and \ref{fig:atlas}(c)). The resulting topographic representations are truncated at 50\% of the peak value to obtain muscle contour for the full atlas (Fig. \ref{fig:atlas}(d)). Truncating the muscle maps allows including only recurrent areas and discarding deviations that result from image misalignments.

Binary classification results from AdaBoost are aligned with the atlas of the respective cross-validation iteration using the procedure described above, considering the bone centroid and the convex hull of the AdaBoost output foreground. The bone is roughly segmented through histogram thresholding. Pixels identified as muscle are then labeled, according to the corresponding pixel in the atlas, to obtain the specific muscle identification. Image transformations are then reversed, which yields a final result in the test MRI frame.

\begin{figure}[t!]
	\begin{minipage}[b]{1\linewidth}
		\centering
		\includegraphics[width=8.5cm]{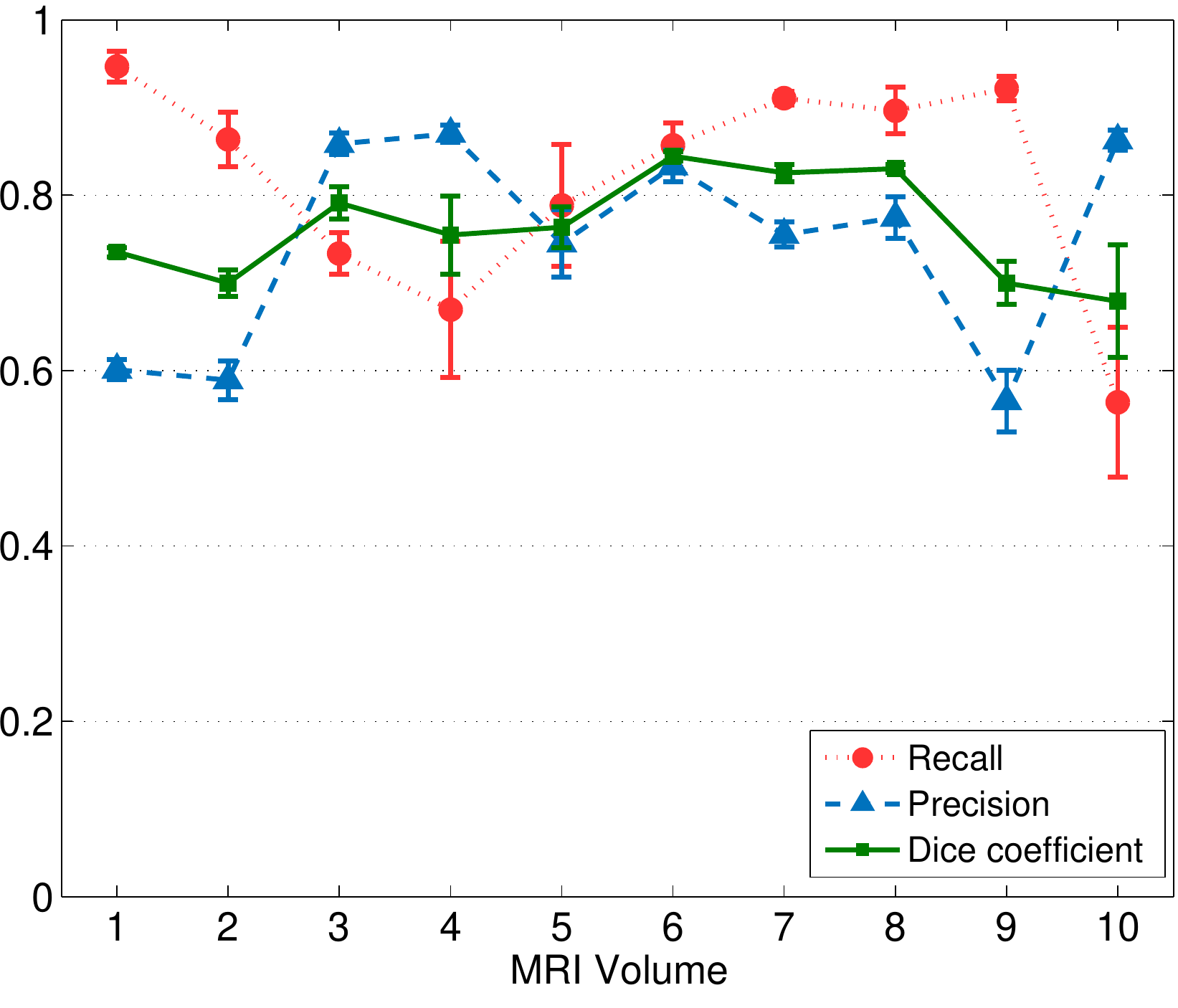}
		\caption{Mean and standard deviation of performance measures per MRI volume.}
		\label{fig:plotrpd}
	\end{minipage}
\end{figure}

\section{Results and Discussion}
\label{sec:results}

The AdaBoost segmentation performance was assessed using recall and precision rates and the Dice overlap coefficient, which were computed, respectively, as follows:

\begin{displaymath}
Recall = \frac{TP}{TP+FN} \eqno{(1)}
\end{displaymath}

\begin{displaymath}
Precision = \frac{TP}{TP+FP} \eqno{(2)}
\end{displaymath}

\begin{displaymath}
Dice = \frac{2|A \cap B|}{|A|+|B|} \eqno{(3)}
\end{displaymath}

In equations (1) and (2), TP refers to true positives, i.e., regions correctly classified as muscle. FN and FP refer to false negatives and false positives, i.e., regions incorrectly classified as non-muscle and muscle, respectively. Equation (3) considers muscle-labeled regions in the segmentation output and ground truth segmentation (A and B).

\begin{figure}[t!]
	\begin{center}
		\subfigure[]{
			\label{result1}
			\includegraphics[width=0.46\linewidth]{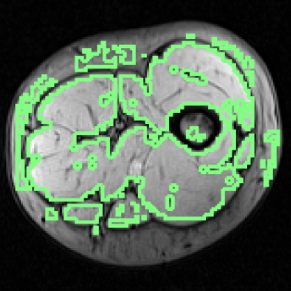}
		}
		\subfigure[]{
			\label{result2}
			\includegraphics[width=0.46\linewidth]{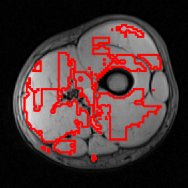}
		}
		\caption{Examples of AdaBoost muscle tissue segmentation results.}
		\label{resultsSegm} \vspace*{-0.4cm}
	\end{center}
\end{figure}

Mean and standard deviation of the mentioned performance measures were taken over a 10-fold cross-validation, and are presented in Fig. \ref{fig:plotrpd}. Segmentation of the whole muscle region (Fig. 4) yielded mean Recall, Precision and Dice overlap coefficients of $0.8150$, $0.7454$ and $0.7623$, respectively. 

These reported results indicate the method efficiency, particularly for the retrieval of true positive occurrences. It should be noted that the AdaBoost algorithm was trained with a \emph{downsampled} ground truth mask to address the descriptor resolution. However, performance was measured considering the normal resolution ground truth mask. 

Average performance measures scored above $0.7$, despite the variability on the results (Fig. \ref{fig:plotrpd}). From the total 50 AdaBoost results obtained in cross-validation, the majority presented a correct identification of the muscle region, with proper tissue separation. Results with high recall rates tend to also increase the number of false positives (lower precision), as the example shown in Fig. 4(a). Using more features could improve the discriminative properties of the proposed classification scheme.

On the other hand, in some cases, the texture recognition failed to provide accurate results, as shown in Fig. 4(b), leading to low recall values. This figure shows the worst obtained result (corresponding to volume 10 in Fig. \ref{fig:plotrpd}). An objective quality study may help to bring insight into a possible impact of texture quality on the recall rate of the proposed segmentation model.

\begin{figure}[t!]
	\begin{center}
	\subfigure[]{
			\includegraphics[width=0.46\linewidth]{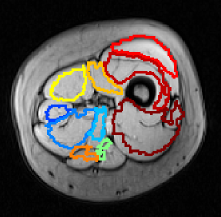}
		}
	\subfigure[]{
			\includegraphics[width=0.46\linewidth]{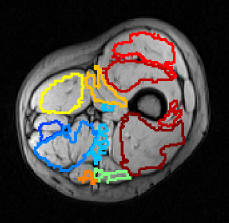}
		}\\
	\subfigure[]{
			\includegraphics[width=0.46\linewidth]{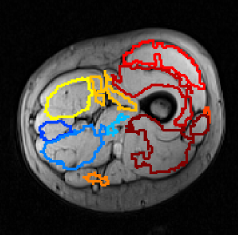}
		}
	\subfigure[]{
			\includegraphics[width=0.46\linewidth]{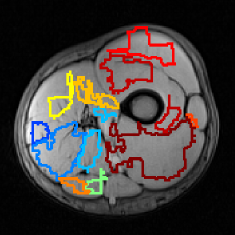}
		}
	\caption{Labeled skeletal muscle segmentation results.}
		\label{fig:resimgs}
	\end{center}
\end{figure}

There is a codependency between AdaBoost accuracy and the atlas transformation and placement. Using the proposed atlas approach to label the AdaBoost binary leads to reasonable results in segmented images with a low false positive rate. However, if false positives have an influence on the position of the most distal point relative to the bone centroid, as shown in the example in Fig. 4(a), the proposed method tends to produce an unreliable muscle labeling. In these cases, the atlas transformation is biased and leads to labels offset, even though the AdaBoost segmentation recall rate is high (Figs. \ref{fig:resimgs}(a) and (b)).

When the recall rate is lower but precision in increased, muscle labels are, generally, placed correctly, despite the total segmented muscle area being slightly affected (Figs. \ref{fig:resimgs}(c) and (d)). Muscles with a smaller cross-sectional area in most of the used MRI, such as the \emph{Sartorius} or the \emph{Rectus femoris}, tend to be unrecognized on the labeled segmentation results if true positive recognition decays in those regions.

Despite the disadvantages of this method, the obtained results suggest that it may perform well in a variety of different scenarios, particularly in cases with more intermuscular tissue or near articulations, where the muscle region of interest is more disperse. Nevertheless, it should be noted that an adequate selection of reference MRI for the atlas construction is critical for a consistent accuracy of this approach, given the variations on muscle geometry and relative positioning, which is quite stable in consecutive slices. 

\section{Conclusion}
The results of this research show that texture features may contribute for a global and automated skeletal muscle segmentation tool, which would greatly improve the state-of-the-art in analysis and diagnosis based on muscle MRI.

Furthermore, this research raises a question on MRI quality and its correlation with texture recognition. Future studies will be conducted to assess the impact of texture quality in this framework. An initial approach to the quality of recognition assessment is already under development.

\section*{Acknowledgment}
The authors would like to acknowledge Dr. Pierre Carlier and his research team at the NMR Laboratory of the Institut de Myologie, Paris, for providing the muscle MRI database.

\bibliographystyle{IEEEbib}
\bibliography{refsMRI}

\begin{thebibliography}{10}

\bibitem{Ahmad2014}
Ezak Ahmad, Moi~Hoon Yap, Hans Degens, and Jamie~S McPhee,
\newblock ``{Atlas-registration based image segmentation of MRI human thigh
  muscles in 3D space},''
\newblock in {\em SPIE Medical Imaging}. International Society for Optics and
  Photonics, 2014, vol. 9037, pp. 90371L--90371L.

\bibitem{Sdika2016}
Michael Sdika, Anne Tonson, Yann Le~Fur, Patrick~J Cozzone, and David Bendahan,
\newblock ``{Multi-atlas-based fully automatic segmentation of individual
  muscles in rat leg},''
\newblock {\em Magnetic Resonance Materials in Physics, Biology and Medicine},
  vol. 29, no. 2, pp. 223--235, 2016.

\bibitem{Troter2016}
Arnaud Le~Troter, Alexandre Four{\'e}, Maxime Guye, Sylviane Confort-Gouny,
  Jean-Pierre Mattei, Julien Gondin, Emmanuelle Salort-Campana, and David
  Bendahan,
\newblock ``{Volume measurements of individual muscles in human quadriceps
  femoris using atlas-based segmentation approaches},''
\newblock {\em Magnetic Resonance Materials in Physics, Biology and Medicine},
  vol. 29, no. 2, pp. 245--257, 2016.

\bibitem{karlsson2015automatic}
Anette Karlsson, Johannes Rosander, Thobias Romu, Joakim Tallberg, Anders
  Gr{\"o}nqvist, Magnus Borga, and Olof Dahlqvist~Leinhard,
\newblock ``{Automatic and quantitative assessment of regional muscle volume by
  multi-atlas segmentation using whole-body water--fat MRI},''
\newblock {\em Journal of Magnetic Resonance Imaging}, vol. 41, no. 6, pp.
  1558--1569, 2015.

\bibitem{orgiu2016automatic}
Sara Orgiu, Claudio~L Lafortuna, Fabio Rastelli, Marcello Cadioli, Andrea
  Falini, and Giovanna Rizzo,
\newblock ``{Automatic muscle and fat segmentation in the thigh from
  T1-Weighted MRI},''
\newblock {\em Journal of Magnetic Resonance Imaging}, vol. 43, no. 3, pp.
  601--610, 2016.

\bibitem{rodrigues2015two}
Rafael Rodrigues, Rui Braz, Manuela Pereira, Jos{\'e} Moutinho, and Antonio~MG
  Pinheiro,
\newblock ``{A two-step segmentation method for breast ultrasound masses based
  on multi-resolution analysis},''
\newblock {\em Ultrasound in Medicine and Biology}, vol. 41, no. 6, pp.
  1737--1748, 2015.

\bibitem{Baudin2012}
P-Y Baudin, Noura Azzabou, Pierre~G Carlier, and Nikos Paragios,
\newblock ``{Prior knowledge, random walks and human skeletal muscle
  segmentation},''
\newblock in {\em International Conference on Medical Image Computing and
  Computer-Assisted Intervention}. Springer, 2012, pp. 569--576.

\bibitem{Andrews2015}
Shawn Andrews and Ghassan Hamarneh,
\newblock ``{The generalized log-ratio transformation: learning shape and
  adjacency priors for simultaneous thigh muscle segmentation},''
\newblock {\em IEEE transactions on medical imaging}, vol. 34, no. 9, pp.
  1773--1787, 2015.

\bibitem{HOG2005}
Navneet Dalal and Bill Triggs,
\newblock ``{Histograms of oriented gradients for human detection},''
\newblock in {\em Computer Vision and Pattern Recognition, 2005. CVPR 2005.
  IEEE Computer Society Conference on}. IEEE, 2005, vol.~1, pp. 886--893.

\bibitem{LOG2009}
Sos Agaian and Ali Almuntashri,
\newblock ``{Noise-resilient edge detection algorithm for brain MRI images},''
\newblock in {\em Engineering in Medicine and Biology Society, 2009. EMBC 2009.
  Annual International Conference of the IEEE}. IEEE, 2009, pp. 3689--3692.

\bibitem{Zhang2011}
Yudong Zhang, Zhengchao Dong, Lenan Wu, and Shuihua Wang,
\newblock ``{A hybrid method for MRI brain image classification},''
\newblock {\em Expert Systems with Applications}, vol. 38, no. 8, pp.
  10049--10053, 2011.

\bibitem{EL2010}
El-Sayed~Ahmed El-Dahshan, Tamer Hosny, and Abdel-Badeeh~M Salem,
\newblock ``{Hybrid intelligent techniques for MRI brain images
  classification},''
\newblock {\em Digital Signal Processing}, vol. 20, no. 2, pp. 433--441, 2010.

\bibitem{vedaldi08vlfeat}
A.~Vedaldi and B.~Fulkerson,
\newblock ``{VLFeat}: An open and portable library of computer vision
  algorithms,'' \url{http://www.vlfeat.org/}, 2008.

\bibitem{Ada1996}
Yoav Freund, Robert~E Schapire, et~al.,
\newblock ``{Experiments with a new boosting algorithm},''
\newblock in {\em Icml}, 1996, vol.~96, pp. 148--156.

\bibitem{adaviola}
Paul Viola, Michael~J Jones, and Daniel Snow,
\newblock ``{Detecting pedestrians using patterns of motion and appearance},''
\newblock {\em International Journal of Computer Vision}, vol. 63, no. 2, pp.
  153--161, 2005.

\end{thebibliography}

\end{document}